\relax
\documentclass[letterpaper]{article} 
\usepackage{aaai19}  
\usepackage{times}  
\usepackage{helvet} 
\usepackage{courier}  
\usepackage[hyphens]{url}  
\usepackage{graphicx} 
\usepackage{graphicx}  
\usepackage[utf8]{inputenc}
\usepackage{hyperref}
\usepackage{booktabs}
\usepackage{graphicx}
\usepackage{xcolor}
\usepackage{amssymb}
\usepackage{subcaption}
\usepackage{amsmath}
\usepackage{xcolor,colortbl}

\definecolor{Gray}{gray}{0.85}

\definecolor{Red}{rgb}{1,0.8,0.8}
\definecolor{Orange}{rgb}{1,1,0.85}
\definecolor{Green}{rgb}{0.8,1,0.8}
\definecolor{LightCyan}{rgb}{0.8,1,1}
\definecolor{Cyan}{rgb}{0.8,1,1}
\definecolor{White}{gray}{1}

\newcolumntype{a}{>{\columncolor{Gray}}c}
\newcolumntype{b}{>{\columncolor{white}}c}

\newcommand{\ourMethodLong}{\textit{Disaggregation via Gaussian Regression} } 
\newcommand{\ourMethod}{\textit{DoGR}}

\title{DoGR: Disaggregated Gaussian Regression for Reproducible Analysis of Heterogeneous Data}
\author{Nazanin Alipourfard,
Keith Burghardt 
\and
Kristina Lerman\\
USC Information Sciences Institute\\
Marina del Rey, CA 90292}

\begin{document}
\maketitle

\begin{abstract}
Quantitative analysis of large-scale data is often complicated by the presence of diverse subgroups, which reduce the accuracy of inferences they make on held-out data. To address the challenge of heterogeneous data analysis, we introduce DoGR, a method that discovers latent confounders by simultaneously partitioning the data into overlapping clusters (disaggregation) and modeling the behavior within them (regression).
When applied to real-world data, our method discovers meaningful clusters and their characteristic behaviors, thus giving insight into group differences and their impact on the outcome of interest. By accounting for latent confounders, our framework facilitates exploratory analysis of noisy, heterogeneous data and can be used to learn predictive models that better generalize to new data. 
We provide the code to enable others to use DoGR within their data analytic workflows \footnote{\url{http://github.com/ninoch/DoGR}}.

\end{abstract}


\section{Introduction} 
\begin{figure}[t!]
    \centering
    \begin{subfigure}[t]{0.8\columnwidth}
        \centering
        \includegraphics[width=\columnwidth]{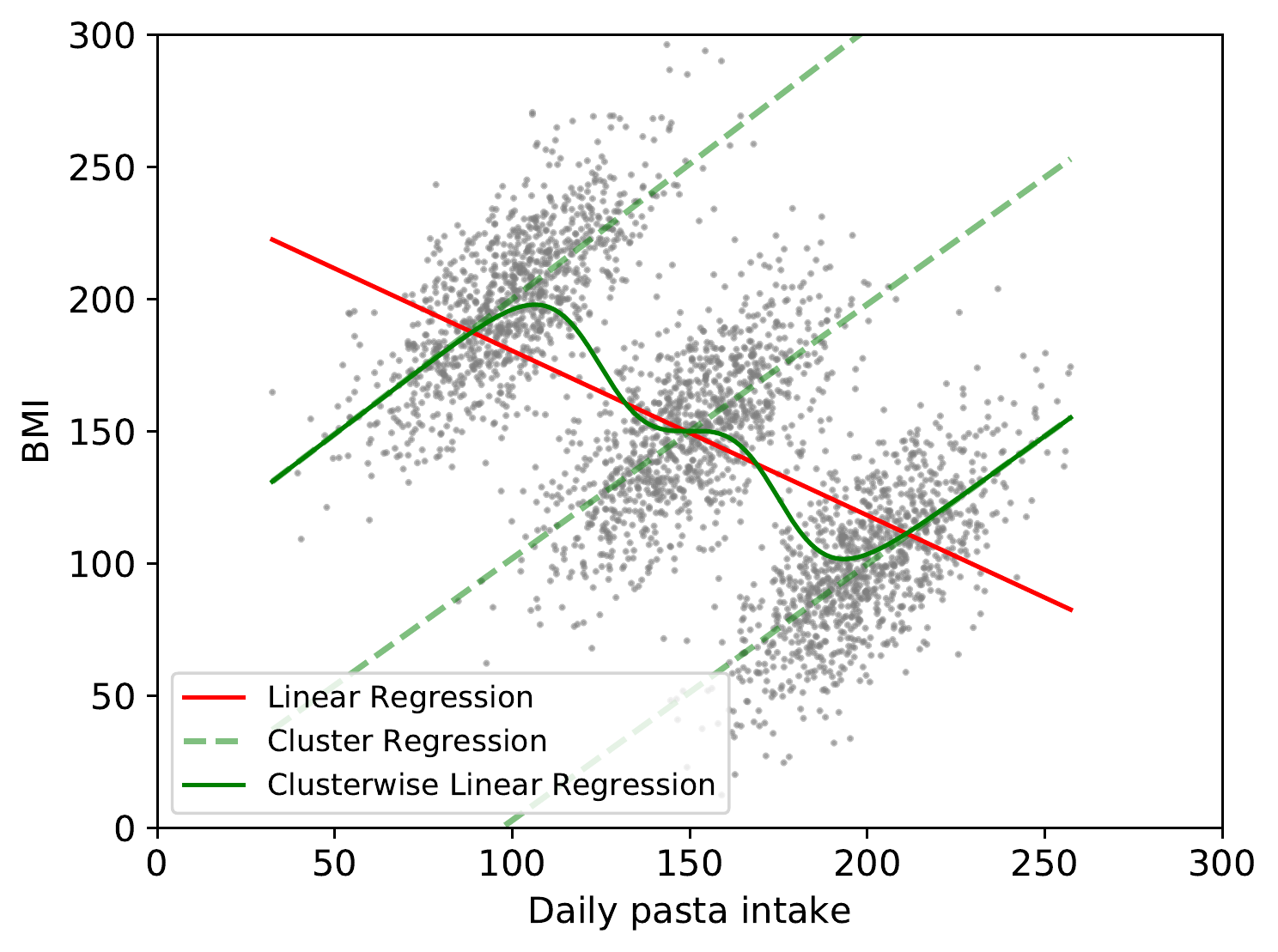}
    \end{subfigure}
    \caption{Heterogeneous data with three latent classes. 
   The figure illustrates Simpson's paradox, where a negative relationship between the outcome and the independent variable exists for population as a whole (dotted line) but reverses when the data is disaggregated by classes (dashed lines).
    \label{fig:explain}}
\end{figure}

Social data is often highly heterogeneous, coming from a population composed of diverse classes of individuals, each with their own characteristics and behaviors. As a result of heterogeneity, a model learned on population data may not make accurate predictions on held-out test data, nor offer analytic insights into the underlying behaviors that motivate interventions. 
To illustrate, consider Figure~\ref{fig:explain}, which shows data collected for a hypothetical nutrition study measuring how the outcome, body mass index (BMI), changes as a  function of daily pasta calorie intake. Multivariate linear regression (MLR) analysis finds a negative relationship in the  population (red dotted line) between these variables. The negative trend suggests that---paradoxically---increased pasta consumption is associated with lower BMI. However, unbeknownst to researchers, the hypothetical population is heterogeneous, composed of classes that varied in their fitness level. These classes (clusters in Fig.~\ref{fig:explain}) represent, respectively, people who do not exercise, people with normal activity level, and athletes. When data is disaggregated by fitness level, the trends within each subgroup are positive (dashed lines), leading to the conclusion that increased pasta consumption is in fact associated with a higher BMI. Recommendations for pasta consumption arising from the naive analysis are opposite to those arising from a more careful analysis that accounts for the confounding effect of different classes of people. The trend reversal is an example  Simpson's paradox, which has been widely observed in many domains, including biology, psychology,  astronomy, and computational social science ~\cite{Chuang2009,kievit2013simpson,minchev2019yule,simpsonsparadox}.  

Social scientists analyze heterogeneous data with underlying structure (classes) using mixed effects models~\cite{winter2013very}. This variant of linear regression allows for intercepts to be random variables in order to account for shifts between classes, and slopes to be random variables in order to model differences in regression coefficients within classes. Mixed effects models are used to describe non-independent observations of data from the same class, and they can even handle trend reversal associated with Simpson's paradox. Mixed effects models assume that classes are specified by a categorical variable, which can be constructed by disaggregating or binning data on an existing variable~\cite{alipourfard2018wsdm}. In practice, however, these classes may not be known a priori, or may be related to multiple variables. Instead, they must be discovered from data, along with the trends they represent.

Many methods already exist for finding latent classes within data; however, they have various shortcomings. Unsupervised clustering methods disaggregate data regardless of the outcome variable being explained although distinct outcomes may be best described by different clusterings  of the same data. Recent methods have tackled Simpson's paradox by performing supervised disaggregation of data ~\cite{alipourfard2018icwsm,fabris2000discovering}, but these methods disaggregate data into subgroups using existing features, and thus are not able to capture effects of latent classes (i.e., unobserved confounders). Additionally, they perform ``hard'' clustering, but assigning each data point to a unique group. Instead, a  ``soft'' clustering is more realistic as it captures the degree of uncertainty about which group the data belong to.
To address these challenges, we describe {\ourMethodLong} ({\ourMethod}), a method that jointly partitions data into overlapping clusters and estimates linear trends within them. This allows for learning accurate and generalizable models
while keeping the interpretability advantage of linear models. Proposed method assumes that the data can be described as a superposition of clusters, or components, with every data point having some probability to belong to any of the components. Each component represents a latent class, and the soft membership represents uncertainty about which class a data point belongs to. We use an expectation maximization (EM) algorithm to estimate component parameters and component membership parameters for the individual data points. {\ourMethod} jointly updates component parameters and its regression coefficients, weighing the contribution of each data point to the component regression coefficients by its membership parameter. The joint learning of clusters and their regressions allows for discovering the latent classes that explain the structure of data. 

Our framework learns predictive and interpretable models of data. {\ourMethod} is validated on synthetic and real-world data and is able to discover hidden subgroups in analytically challenging scenarios, where subgroups overlap along some of the dimensions, and identify interesting trends that may be obscured by Simpson's paradox. This includes scenarios where the trends in data show Simpson's reversal.
Our method achieves performance on prediction tasks comparable to or better than state-of-the-art algorithms, but in a fraction of the time, making it well suited for big data analysis.

\section{Background}
\label{sec-related}
Regression is one of the most widely used methods in data analysis due to its flexibility, interpretability, and computational scalability \cite{Fox1997}. Linear models are perhaps one of the most popular regression methods to learn the relationship between predictor variables (also known as fixed effects or features) and the outcome variable. However, linear models make assumptions about the data that are often violated by real-world data. One of these assumptions is independence of observations. In reality, however, observations may come from groups of similar individuals, violating the independence assumption. To account for individual differences within such groups, mixed-effect or multi-level models were developed \cite{McLean1991}. 
These models assume that similar individuals, or data samples, come from the same class, and that classes are specified by some categorical variable~\cite{winter2013very}. In some cases, this variable can be created by binning a continuous variable and  disaggregating data by these  bins~\cite{alipourfard2018wsdm}. 
However, existing methods do not work when differences between groups or individuals cannot directly be described by observed variables or features. In this chapter, we describe a method that captured differences between groups as latent confounders (i.e., latent clusters) that are learned from data. Our work improves the utility of linear models in the analysis of social data by controlling for these clusters. 
A number of previous works have attempted to tackle the issue of latent confounders. Clusterwise linear regression (CLR) \cite{spath} starts with initial clusters and updates them by reassigning one data point to another partition that minimizes a loss function. The method is slow, since it moves one data point at each iteration. Two other methods, WCLR~\cite{silva2017} and FWCLR~\cite{silva2018}, improve on CLR by using k-means as their clustering method. These methods were shown to outperform CLR and other methods, such as K-plane \cite{manwani2015}. 
 In Conditional Linear Regression \cite{calderon2018}, the goal is to find a linear rule capable of achieving more accurate predictions for a \textit{segment} of the population by ignoring outliers. One of the larger differences between their method and ours is that Conditional Linear Regression focuses on a small subset of data, while we model data as a whole, alike to CLR, WCLR, FWCLR and K-plane. 
The shared parameter $\lambda$ across all clusters in WCLR and FWCLR makes these methods perform poorly if clusters have different variance, as we will show in our results section. 
Other methods have combined Gaussian Mixture Models and regression to create algorithms similar to our own \cite{sung2004,ghahramani1994}. We call these methods \textit{Guassian Mixture Regression (GMR)}. In contrast to these methods, however, we can capture relationships between independent and outcome variables through regression coefficients, which previous methods were unable to do. We also use \textit{Weighted Least Squares} to fit our model, which makes it less sensitive to outliers \cite{Cleveland1988}. 

Causal inference has been used to address the  problem of confounders, including latent confounders \cite{Louizos2017}, and to infer the causal relationship between features and outcomes \cite{Bareinboim2016,Ranganath2018,Wang2018}. One difficulty with causal inference however, is that the focus is traditionally on one intervention \cite{Ranganath2018}. Taking into account synergistic effects of multiple features is not well understood, but has been attempted recently \cite{Ranganath2018,Wang2018}. With adequate datasets, these can help infer causal relationships between multiple causes, but certain causal assumptions are needed, which might not correspond to reality. In contrast, regression offers us the opportunity to understand relationships between each feature and an outcome, regardless of the dataset, even if we cannot make causal claims.
\section{{\ourMethod} Method}

Let $\mathcal{D} = \{d_1, d_2, d_3, ..., d_N | d_i = (x_i, y_i) \in \mathbb{R}^p \times \mathbb{R}\}$ be a set of $N$ observations, or  records, each containing  a real-valued outcome $y_i$ of an observation $i$ and a vector of $p$ independent variables, or features, $[x_{i, 1}, x_{i, 2}, ..., x_{i, p}]$. Regression analysis is often used to capture the relationship between each independent variable and the outcome. Specifically, MLR estimates regression coefficients $\beta_0, \beta_1, ..., \beta_p$ by minimizing the residuals of $y = \beta_0 + \beta_1 x_1 + \beta_2 x_2 + ... + \beta_p x_p$ over all observations. However, parameters learned by the \textit{MLR} model may not generalize to out-of-sample populations, as they can be confounded by Simpson's paradox and sampling bias. We can call this the ``robustness'' problem of regression. 



As discussed in the introduction, Figure \ref{fig:explain}(a) illustrates this problem with synthetic data. The \textit{MLR} model trained on the aggregate  data  gives $BMI = 241 - 0.61 x$. This suggests a negative relationship (solid red line) between the independent variable $x=$\textit{Daily pasta intake} and the outcome $BMI$. However, there are three components each with a positive trend. Indeed, applied separately to each component, \textit{MLR} learns the proper positive relationship (dashed green line) between \textit{Daily pasta intake} and $BMI$: $BMI = \beta_0 + \beta_1 x$, where $\beta_0$ is cluster's intercept, $\{-108, -13, 102\}$ and $\beta_1$ is coefficient, $\{1.03, 1.08, 0.97\}$. Associations learned by the \textit{MLR} model trained on the population-level data are not \textit{robust} or \textit{generalizable}. This could help explain the \emph{reproducibility problem} seen in many fields \cite{Pashler2012,Smaldino2016}, where trends seen in some experiments cannot be reproduced in other experiments with different populations. This can be clearly observed in Figure \ref{fig:explain}(b), which represents a different sampling of the population. The underlying clusters are the same, but the middle cluster is over-sampled during data collection. As a result, \textit{MLR} trained at the population-level finds only a weak association between independent variable and the outcome (solid red line). In contrast, regressions learned separately for each cluster (green dashed line) remain the same even in the new data: thus, the cluster regressions represent \textit{robust} and \textit{generalizable} relationships in data.

\subsection{Model Specification}
The goal of this work is to learn robust and reproducible trends through a regression model that accounts for the  latent structure of data, for example, the presence of three clusters in data shown in Fig.~\ref{fig:explain}. Our model jointly \textit{disaggregates} the data into $K$ overlapping subgroups, or components, and performs \textit{weighted linear regression} within each component.
We allow components to overlap in order to represent the uncertainty about which component or subgroup an observation belongs to. 

In what follows,  we used  capital letters to denote {random variables} and lowercase letters their values. 
We model the independent variable $X$ of each component $k$ as a multivariate normal distribution with mean $\mu_k \in \mathbb{R}^p$ and covariance matrix $\Sigma_k \in \mathbb{R}^{p\times p}$: 
\begin{equation}
f^{(k)}_{X} \sim \mathcal{N}(\mu_k, \Sigma_k)
\label{eq:marginal}
\end{equation}
In addition, each component is characterized by a set of regression coefficients $\mathbf{\beta_k} \in \mathbb{R}^{p+1}$. The regression values of the component $k$ are 
\begin{equation}
\label{eq:y-hat}
\hat{Y}^{(k)} = \beta_{k, 0} + \beta_{k, 1} X_{1} + \beta_{k, 2} X_{2} + ... + \beta_{k, p} X_{p},     
\end{equation}
with $y - \hat{y}^{(k)}$ giving the residuals for component $k$. Under the assumption of normality of residuals~\cite{desarbo1988maximum}, $Y$ has a normal distribution with mean $\hat{Y}^{(k)}$ and standard deviation $\sigma_k$.

\begin{equation}
f^{(k)}_{Y | X} \sim \mathcal{N}(\hat{Y}^{(k)}, \sigma_k),
\label{eq:conditional}
\end{equation}

\noindent where $\hat{Y}^{(k)}$ 
is defined by Eq.~\ref{eq:y-hat}.
Under the assumption of \textit{homoscedasticity}, in which the error is the same for all $X$, the joint density is the product of the conditional (Eq.~\ref{eq:conditional}) and marginal (Eq.~\ref{eq:marginal}) densities: 

\begin{equation} \label{eq:joint}
\begin{aligned}
    f^{(k)}_{X, Y}(x, y) = f_{Y | X}(y|x) f_{X}(x) \\
    = \varphi(y; \hat{y}^{(k)}, \sigma_k) \varphi(x; \mu_k, \Sigma_k)
\end{aligned}
\end{equation}

Multiplication of $f_{Y | X}(y|x) f_{X}(x)$ can be converted to a normal distribution: $f^{(k)}_{X, Y}(x, y)$ comes from $\mathcal{N}(\mu'_k, \Sigma'_k)$, where $\mu'_k = [\mu_k^{(1)}, \mu_k^{(2)}, \ldots, \mu_k^{(d)}, \hat{y}^{(k)}]$ and $\Sigma'_k = \left[{\begin{array}{cc} \Sigma_k & 0 \\ 0 & \sigma_k \\ \end{array} } \right]$ is a block matrix.  


\subsection{Model Learning} 

\noindent The final goal of the model is to predict the outcome, given independent variables. This prediction combines the predicted values of outcome from all components by taking the average of the predicted values, weighed by the size of the component. We define $\omega_k$ as the weight of the component $k$, where $\sum_{k} \omega_k = 1$. We can define the joint distribution over all components as $f_{X, Y}(x, y) = \sum_{k = 1}^{K} \omega_k \times f^{(k)}_{X, Y}(x, y)$.

\noindent Then the \textit{log-likelihood} of the model over all data points is:

\begin{equation}
    \mathcal{L} = \sum_{i=1}^{N} log \big(\sum_{k=1}^K \omega_k \times f^{(k)}_{X, Y}(x_i, y_i)\big)
    \label{fig:LL}
\end{equation}

\noindent The formula here is same as the \textit{Gaussian Mixture Model}, except that the target $y_i$ is a function of $x_i$. To find the best values for parameters $\theta = \{\omega_k, \mu_k, \Sigma_k, \beta_k, \sigma_k | 1 \leq k \leq K \}$ we can leverage the \textit{Expectation Maximization} (EM) algorithm. The algorithm iteratively refines parameters based on the expectation (E) and maximization (M) steps.

\subsubsection{\textbf{E-step}}
\noindent Let's define $\gamma_{i, k}$ (\textit{membership parameter}) as
probability that data point $i$ belongs to component $k$. Given the parameters $\theta_t$ of last iteration, membership parameter is:

\begin{equation}
    \gamma_{i, k} = \frac{\omega_k \times f^{(k)}_{X, Y}(x_i, y_i)}{\sum_{k'} \omega_k' \times f^{(k')}_{X, Y}(x_i, y_i)}
\end{equation}

\noindent Thus, the E-step disaggregates the data into clusters, but it does so in a ``soft'' way, with each data point having some probability to belong to each cluster. 

\subsubsection{\textbf{M-step}} Given the updated membership parameters, the M-step updates the parameters of the model for the next iteration as $\theta_{t+1}$:
$$\omega_k = \frac{\sum_i \gamma_{i, k}}{N}$$
$$\mu_k = \frac{\sum_i \gamma_{i, k} x_i }{\sum_i \gamma_{i, k}}$$
$$\Sigma_k = \frac{\sum_i \gamma_{i, k} (x_i - \mu_k)(x_i - \mu_k)^T}{\sum_i \gamma_{i, k}}$$

\noindent In addition, this step updates regression parameters based on the estimated parameters in $\theta_t$.
Our method uses \textit{Weighted Least Squares} for updating the regression coefficients  $\beta_k$ for each component, with $\gamma$ as weights. In the other words, we find $\beta_k$ that minimizes the \textit{Weighted Sum of Squares (WSS)} of the residuals:
$$WSS(\beta_k) = \sum_{i} \gamma_{i, k} (y_i - (\beta_{k, 0} + \beta_{k, 1} x_{i, 1} + ... + \beta_{k, p} x_{i, p}))^2.$$

\noindent Using the value of $\beta_k$, the updated $\sigma_k$ would be:

$$\sigma_k = \frac{\sum_i \gamma_{i, k} (y_i - \hat{y}_i^{(k)})^2}{\sum_i \gamma_{i, k}}$$

\noindent Intuitively, $\mu_k$ shows us where in $\mathbb{R}^p$ the center of each subgroup $k$ resides. The further a data point is from the center, the lower its probability to  belong to the subgroup. The covariance matrix $\Sigma_k$ captures the spread of the subgroup in the space $\mathbb{R}^p$ relative to the center $\mu_k$. Regression coefficient $\beta_k$ gives low weight to outliers (i.e., is a \emph{weighted} regression) and captures the relationship between $X$ and $Y$ near the center of the subgroup. Parameter $\sigma_k$ tells us about the variance of the residuals over fitted regression line, and $\omega_k$ tells us about the importance of each subgroup. 

\subsection{Prediction}
For test data $x$, the predicted outcome is the weighted average of the predicted outcomes for all components. The weights capture the uncertainty about which component the test data belongs to.
Using equation~\ref{eq:conditional}, the best prediction of outcome for component $k$ would be $\hat{Y}^{(k)}$, which is the mean of \textit{conditional outcome value} of the component $k$: 

\begin{equation}
    \hat{y} = \sum_{k = 1}^K \omega_k \times (\beta_{k, 0} + \beta_{k, 1} x_{1} + ... + \beta_{k, p} x_{p})
\end{equation}
The green solid line in figure~\ref{fig:explain} represents the predicted outcome $\hat{y}$ as function of $x$. The solid green line, is weighted average over dashed-green lines. 

\section{Data}
We apply our method to a synthetic and real-world data sets, including large social data sets described below.

\begin{figure}[t]
    \centering
    \begin{subfigure}[t]{\columnwidth}
        \centering
        \includegraphics[width=0.7\columnwidth]{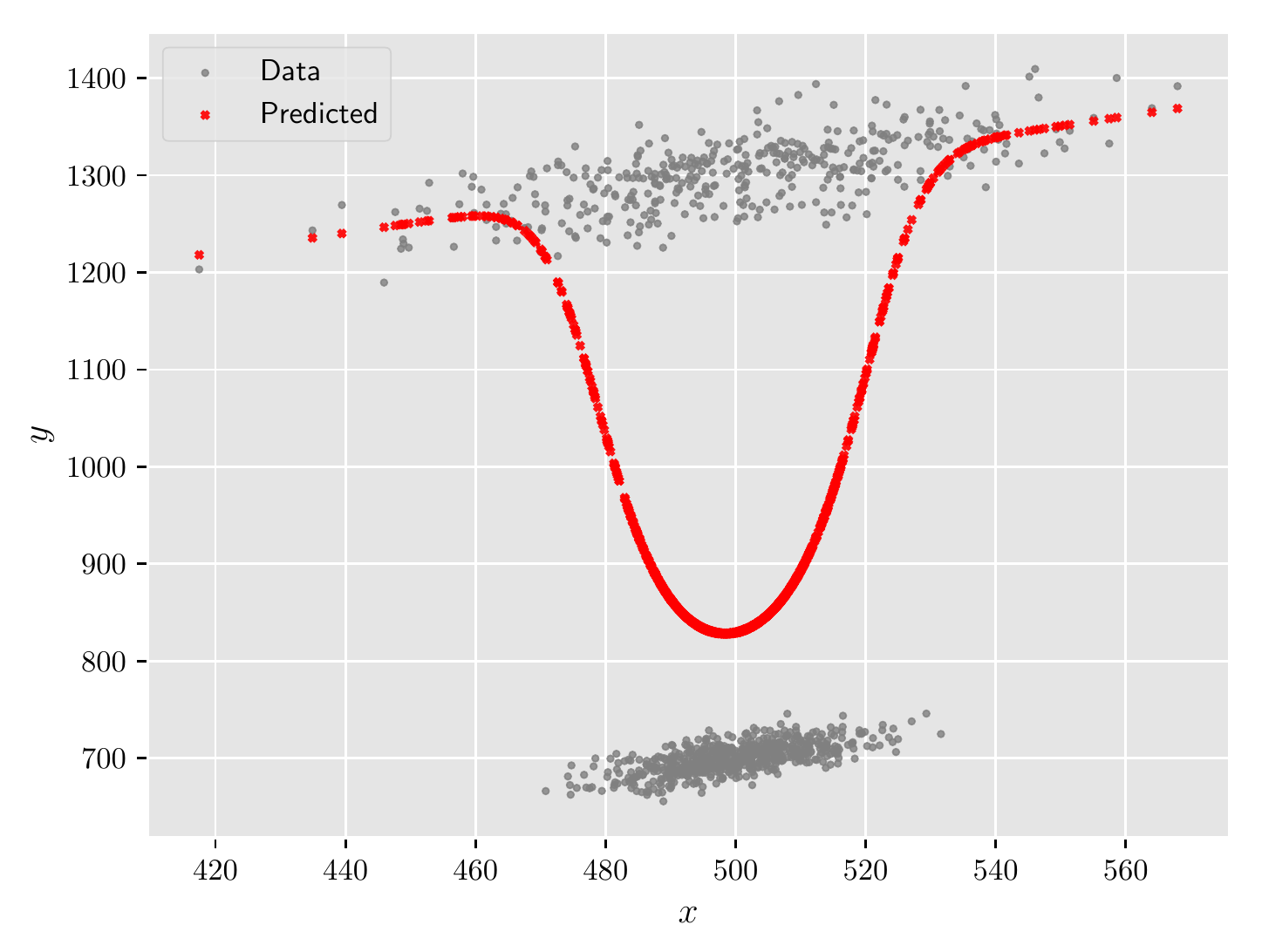} 
    \end{subfigure}
    \caption{Synthetic data with two components centered on $x = 500$, but with different variances. Gray points are data, and red points are predicted outcomes made by our method.   }\label{fig:synthetic}
\end{figure}

The \textbf{Synthetic} data consists of two subgroups, with the same mean on $x$, but different variances, as shown in  Figure~\ref{fig:synthetic}. The variance of $x$ for the top component is $600$ and for the bottom component is $100$. The number of data points in bottom component is $3,000$ and in top component $2,000$. The $y$ value for the bottom component is $y = 200 + x$ and the top component it is $y = 800 + x$, with $20$ as variance of residual.

The \textbf{Metropolitan} data comes from a study of emotions expressed through Twitter messages posted from locations around Los Angeles County~\cite{Arora2016icwsm}. 
a large metropolitan area.
The data contains more than 6 million geo-referenced tweets from 33,000 people linked  to US census tracts through their locations. The demographic and socioeconomic characteristics of tracts came from the 2012 American Fact Finder.\footnote{http://factfinder.census.gov/}
A tract is a small region that contains about 4,000 residents on average. The tracts are designed to be relatively homogeneous with respect to socioeconomic characteristics of the population, such as income, 
residents with bachelors degree or above, and other demographics.  
Of more than 2000 tracts within this metropolitan area, 1688 were linked to tweets. 

Emotional \textit{valence} was estimated from the text of the tweets 
using Warriner, Kuperman, and Brysbaert (WKB) lexicon~\cite{Warriner2013}. The lexicon gives valence scores---between 1 and 9---which quantify the level of pleasure or happiness expressed by a word, for 14,000 English words.  Since valence is a proxy for the expressed happiness, the data allows us to investigate the social and economic correlates of happiness. We use valence as the outcome in data analysis.

The \textbf{Wine Quality} data combines two benchmark data sets from \href{https://archive.ics.uci.edu/ml/datasets/wine+quality}{UCI repository}\footnote{https://archive.ics.uci.edu/ml/datasets/wine+quality} related to red and white wines~\cite{winequality}. The data contains records about $4898$ white wine  and $1599$ red wine samples of wine quality and concentrations of various chemicals. 
\textit{Quality} is a score between $0$ and $10$, and is typically around $5$. Our goal is to model wine quality based on physicochemical tests. Note that the type of wine (red or white) 
is only used to evaluate the learned components. 

The \textbf{New York City Property Sale} data\footnote{https://www.kaggle.com/new-york-city/nyc-property-sales} contains records of every building or building unit (such as an apartment) sold in the New York City property market over a 12-month period, from September 2016 to September 2017. We removed all categorical variables like \textit{neighborhood} and \textit{tax class}. We use each property's \textit{borough} to study relevance of the components to different neighborhoods in NYC; however, we do not use it in analysis. The outcome variable is \textit{sale price}, which is in million dollars, and it ranges from $0$ to $2,210$. 

The \textbf{Stack Overflow} dataset contains data from a question-answering forum on the topic of computer programming. Any user can ask a question, which others may answer. We used anonymized data representing all answers to questions with two or more answers posted on Stack Overflow from August 2008 until September 2014.\footnote{\url{https://archive.org/details/stackexchange}}
We created a user-focused data set by selecting at random one answer written by each user and discarding the rest. 
To understand factors affecting the length of the answer (the outcome, which we measure as the number of \textit{words} in the answer), we use features of answers and features of users. 
Answer features include the number of \emph{hyperlinks}, and \emph{lines of code} the answer contains, and its \emph{Flesch readability} score~\cite{Readability}. Features describing answerers are their \emph{reputation}, \emph{tenure} on Stack Overflow (in terms of \emph{percentile} rank) and experience, i.e., the total \emph{number of answers} written during their tenure. We also use activity-related features, including \emph{time since previous answer} written by the user, \emph{session length}, giving the number of answers user writes during the session, and \emph{answer position} within that session. We define a session as a period of activity without a break of 100 minutes or longer. Features \emph{acceptance probability}, \emph{reputation} and \emph{number of answers} are only used to evaluate the learned components. 

\subsection{Preprocessing}
When variables are linearly correlated, regression coefficients are not unique, which presents a challenge for analysis. To address this challenge, we used Variance Inflation Factor (VIF) to remove multicollinear variables. We iteratively remove variables when their VIF is larger than five. For example, in the Metropolitan data, this approach reduced the number of features from more than 40 to six, representing the number of residents within a tract who are $White$, $Black$, $Hispanic$ and $Asian$, as well as percent of adult residents with $Graduate$ degree or with incomes $Below\ Poverty$ line. Table~\ref{tab:info}, represents information about all data sets. 

\begin{table}[h]
    \centering
    
    \resizebox{0.47\textwidth}{!}{
    \begin{tabular}{c|c|c|c|c}
        \textit{Dataset} & \textit{records} & \textit{features} & \textit{after VIF} & \textit{outcome} \\
        \hline 
        Synthetic & $5,000$ & $1$ & $1$ & Y \\ 
        Metropolitan & $1,677$ & $42$ & $6$ & Valence \\ 
        Wine Quality & $6,497$ & $11$ & $5$ & Quality \\
        NYC & $36,805$ & $7$ & $4$ & Sale Price\\
        Stack Overflow & $372,321$ & $13$ & $5$ & Answer Length \\
        \hline
    \end{tabular}}
    \caption{Data sets and their characteristics. \label{tab:info}}
\end{table}

     


\subsection{Finding Components}
As first step in applying {\ourMethod} to data for qualitative analysis, we need to decide on the number of components. In general, finding the appropriate number of components in clustering algorithms is difficult and generally requires \text{ad hoc} parameters. For Gaussian Mixture Models, the \textit{Bayesian information criterion (BIC)} is typically used. We also use BIC with $k \times (p^2 + 2p + 3)$ parameters, where $k$ is number of components and $p$ is number of independent variables. 
Based on the BIC scores, 
we choose five components for this data. In our explorations, using three or six components gave qualitatively similar results, but with some of the components merging or splitting. 
With the same procedure, the optimal number of components for \textit{Wine Quality} and \textit{NYC} and \textit{Stack Overflow} data is four. 

\section{Results}
Due to the unique interpretability of our method, we first use it to describe meaningful relationships between variables in the real-world data (``Qualitative Results''). We then show how it compares favorably to competing methods (``Quantitative Results'').

\subsection{Qualitative Results}
We use a \textit{radar chart} to visualize the components discovered in the data. Each colored polygon represents mean of the component, $\mu$, in the feature space. Each vertex of the polygon represents a feature, or dimension, with the length giving the coordinate of $\mu$ along that dimension. For the purpose of  visualization, each coordinate value of $\mu$ was divided by the largest coordinate value of $\mu$ across all components. The maximum possible value of each coordinate after normalization is one. The mean value of the outcome variable within each component is shown in the legend, along with 95\% confidence intervals.

Our method also estimates the regression coefficients, which we can compare to MLR. We computed the p-value assuming coefficients in \ourMethod are equal to \textit{MLR}. If $\beta_0$ and $\beta_1$ are two coefficients to compare, and $\sigma_{\bar{\beta_0}}$ and $\sigma_{\bar{\beta_1}}$ are the \textit{standard errors} of each coefficient, then the z-score is
$z = \frac{\beta_{0} - \beta_{1}}{\sqrt{\sigma_{\bar{\beta_0}}^2 + \sigma_{\bar{\beta_1}}^2}}$, from which we can easily infer the p-value assuming a normal distribution of errors \cite{coefficient_test}.

\begin{figure}[t]{}
    \begin{subfigure}[t]{\columnwidth}
        \centering
        
        \includegraphics[width=\columnwidth]{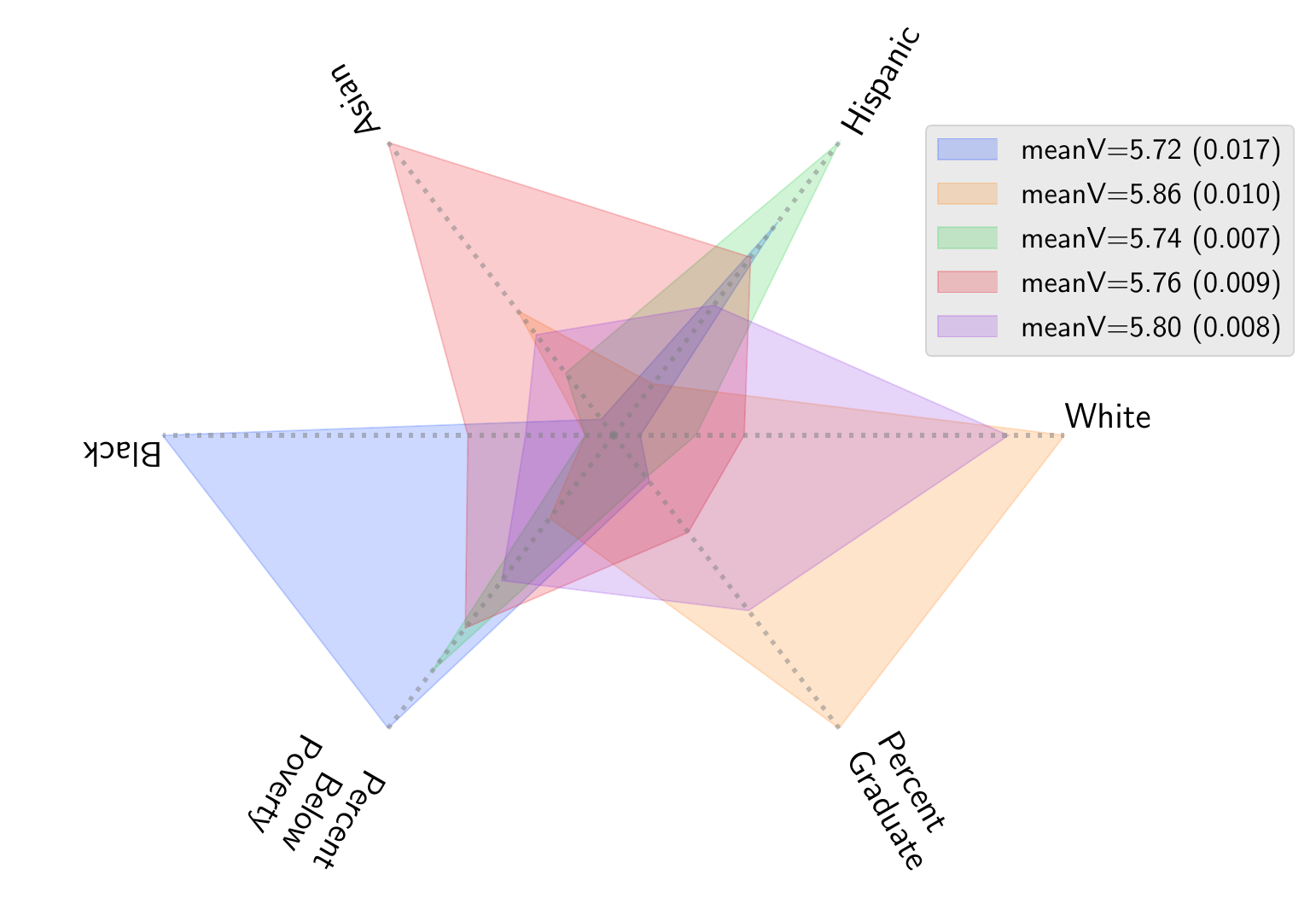}
    \end{subfigure}
    \centering
    \resizebox{0.40\textwidth}{!}{
    \begin{tabular}{|c||c|c||l||l|}
        \cline{4-5}
        \multicolumn{3}{c||}{} &  \multicolumn{1}{|c||}{\textit{\% Below}} & \multicolumn{1}{|c|}{\textit{\%}} \\ 
        \multicolumn{3}{c||}{} &  \multicolumn{1}{|c||}{\textit{Poverty}} & \multicolumn{1}{|c|}{\textit{Graduate}} \\ 
        \cline{2-5}
        \multicolumn{1}{c||}{} & \textit{mean} & \textit{std} & \multicolumn{1}{|c||}{$\beta$} & \multicolumn{1}{|c|}{$\beta$} \\ 
        \hline
        \textit{All} & $5.78$ & $0.098$ & \multicolumn{1}{|c||}{$0.0000$} & \multicolumn{1}{|c|}{$0.0028$} \\ 
        \hline
        \hline
        \textit{Orange} &  $5.86$ & $0.092$ & $0.0063^{***}$ & $0.0016^{*}$ \\ 
        \textit{Purple} & $5.80$ &  $0.077$ & $0.0002$ & $0.0030$ \\ 
        \textit{Red} & $5.76$ &  $0.080$ & $0.0018^{***}$  & $0.0050^{***}$ \\ 
        \textit{Green} &  $5.74$ & $0.073$ & $0.0001$  & $0.0056^{***}$  \\ 
        \textit{Blue}  & $5.72$ & $0.117$ & $-0.0008^{***}$  & $0.0001^{***}$  \\ 
        \hline
        \multicolumn{5}{|l|}{* \footnotesize{p-value$ \le 0.05$}; {*** \footnotesize{p-value$ \le 0.001$}}} \\ \hline
    \end{tabular}}
    \caption{Disaggregation of the Metropolitan data into five subgroups. The radar plot shows the relative importance of a feature within the subgroup. The table report regression coefficients for two independent variables $Percent\ Below\ Poverty$ and $Percent\ Graduate$ for Multivariate Regression (MLR) of aggregate data and separately for each subgroup found by our method. 
\label{fig:LA_mu5}}
\end{figure}

\begin{figure}[h]
    \centering
        \includegraphics[width=1.06\columnwidth,height=10cm]{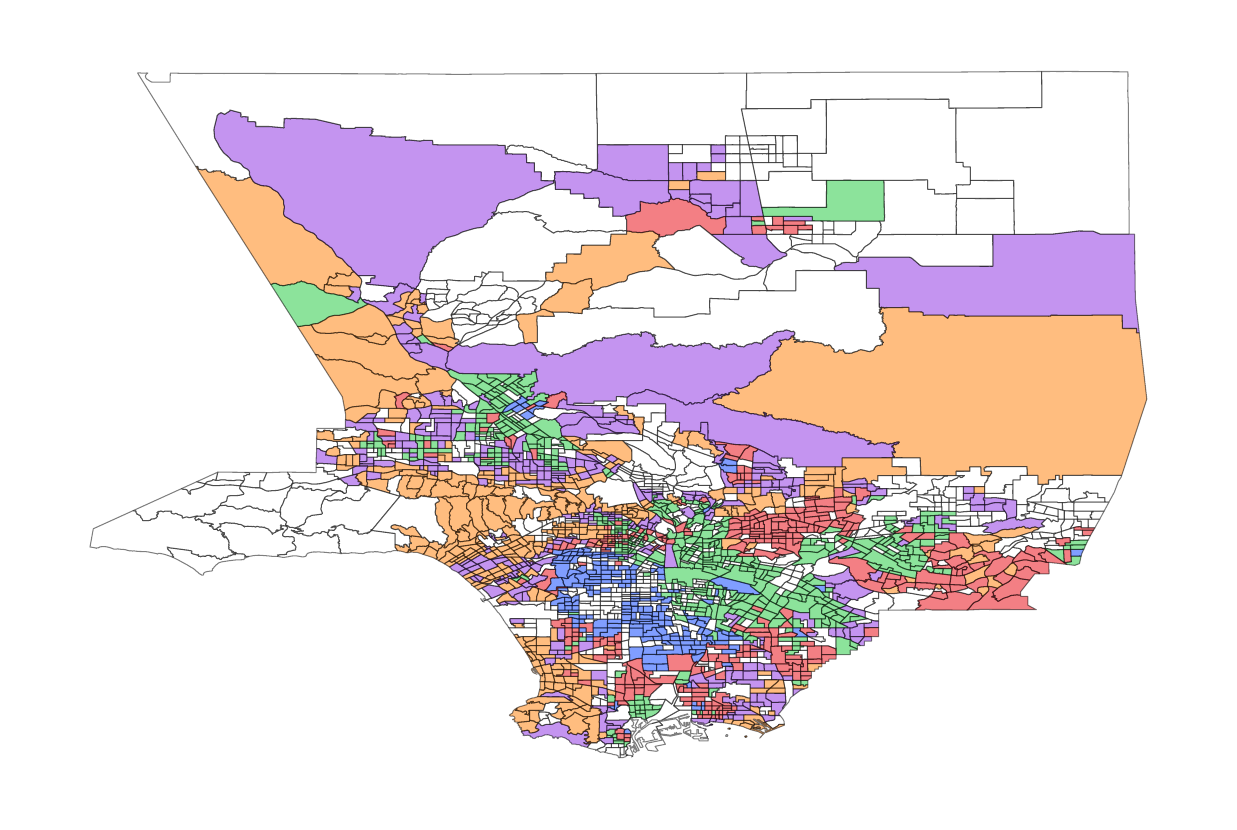} 
     
    \caption{Map of Los Angeles county. The color represent the component for the Tract. Colors match the colors of Figure \ref{fig:LA_mu5}. Pasadena Neighbourhood is red, while downtown is blue, Beverly Hills is orange, Santa Monica is purple and orange, while we can see more orange for Manhattan beach. East Los Angeles is green. \label{fig:map}}
\end{figure}

\subsubsection{\textbf{Metropolitan}}
Figure~\ref{fig:LA_mu5} visualizes the components of the \textit{Metropolitan} data and reports regression coefficients for two variables. 
Interestingly, the data is disaggregated along ethnic lines, perhaps reflecting segregation within the metropolitan area. Figure \ref{fig:map} represents the regions in each component in Los Angeles County map. The $Orange$ component consists of census tracts with many highly educated white residents. It also has highest valence ($5.86$), meaning that people post the happiest tweets from those tracts
. The \textit{Purple} component, is ethnically diverse, with large numbers of white, Asian and Hispanic residents. It's valence ($5.80$) is only slightly lower than of the \textit{Orange} component. The \textit{Red} component has largely Asian and some Hispanic neighborhoods that are less well-off, but with slightly lower valence ($5.76$). The \textit{Blue} and \textit{Green} components represents tracts with the least educated and poorest residents. They are also places with the lowest valence ($5.74$ and $5.72$, respectively). Looking at the regression coefficients, education is positively related to happiness across the entire population (\textit{All}), and individually in all components, with the coefficients significantly different from  \textit{MLR} in four of the five components, \textit{suggesting this trend was a Simpson's paradox}. However, the effect is weakest in the most and least educated components. 
Poverty has no significant effect across the entire population, but has a negative effect on happiness in poorest neighborhoods (\textit{Blue}). Counter-intuitively, regression coefficients are positive for two components. It appears that within these demographic classes, the poorer the neighborhood, the happier the tweets that originate from them. 

\subsubsection{\textbf{Wine Quality}}
\begin{figure}[t]
    \centering
    \begin{subfigure}[t]{0.9\columnwidth}
        \centering
        \includegraphics[width=\columnwidth]{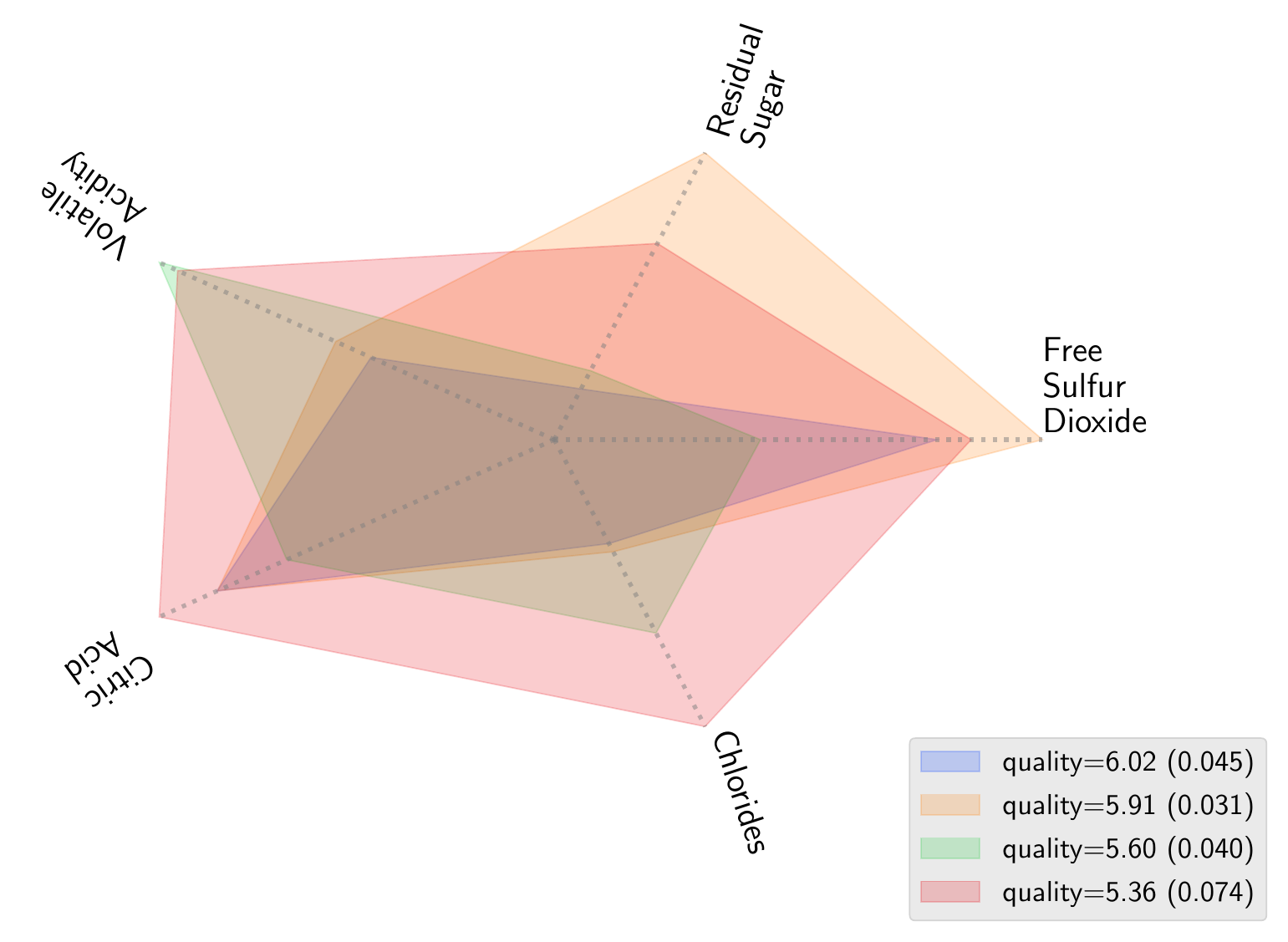} 
    \end{subfigure}
\footnotesize{
\resizebox{0.47\textwidth}{!}{
    \begin{tabular}{|c||c|c||l|l|l|l|}
        \cline{2-6}
        \multicolumn{1}{c|}{} &
        \multicolumn{2}{c||}{\textit{Wine Quality}} & \multicolumn{1}{|c|}{\textit{Citric Acid}} & \multicolumn{1}{|c|}{\textit{SO2}} & \multicolumn{1}{|c|}{\textit{Sugar}}\\ 
        \cline{2-6}
        \multicolumn{1}{c||}{} & \textit{mean} & \textit{std err} & \multicolumn{1}{|c|}{$\beta$} & \multicolumn{1}{|c|}{$\beta$}& \multicolumn{1}{|c|}{$\beta$} \\
        \hline
\textit{All}	&	    	&		&	$0.1044$		&	$-0.0008$		&	$-0.0175$		\\ \hline
\textit{Blue}	&	6.02	&	$0.023$	&	$0.4299$		&	$0.0173^{***}$	&	$0.4272^{***}$	\\
\textit{Orange}	&	5.91	&	0.016	&	$-0.1545^{**}$	&	$-0.0001$		&	$-0.0156$		\\
\textit{Green}	&	5.6	&	0.020	&	$-0.0968$		&	$-0.0024$		&	$0.3227^{***}$	\\
\textit{Red}	&	5.36	&	0.037	&	$0.3039^{**}$	&	$-0.011^{***}$	&	$-0.004^{***}$	\\
\hline
        \multicolumn{6}{|l|}{* \footnotesize{p-value$ \le 0.05$}; {*** \footnotesize{p-value$ \le 0.001$}}} \\ \hline
    \end{tabular}}
}
        
    \caption{The value of the center of each component ($\mu$) for four components of the Wine Quality data. The \textit{Blue} and \textit{Orange} components are almost entirely white wines, and the \textit{Green} component is composed mostly ($85\%$) of red wines. The lowest quality and smallest---\textit{Red}---component is a mixture of red ($43\%$) and white ($57\%$) wines.
    \label{fig:wine_mu}}
\end{figure}

Figure~\ref{fig:wine_mu} visualizes the disaggreagation of the  \textit{Wine Quality} data into four components. 
Although we did not use the type of wine (red or white) as a feature, wines were naturally disaggregated by type. 
The \textit{Blue} component is almost entirely ($98\%$) composed of high quality ($6.02$) white wines. 
All data in the \textit{Orange} component ($5.91$) are white wine, while the  \textit{Green} component is composed mostly ($85\%$) of red wines with average quality $5.60$. The lowest quality (\textit{Red}) component, with quality equal to $5.36$, contains a mixture of red ($43\%$) and white ($57\%$) wine. In other words, we discover that in higher-quality wines, \textit{wine color can be determined with high accuracy simply based on its chemical components}, which was not known before, to the best of our knowledge. Low quality wines appear less distinguishable based on their chemicals. We find \textit{Chlorides} have a negative impact on quality of wine in all components (not shown), \textit{Sugar} has positive impact on high quality white wines and red wines, and negative impact on low quality wines. Surprisingly, \textit{Free Sulfur Dioxide} has a positive impact on high quality white wines, but a negative impact in other components. These are findings that may be important to wine growers, and capture subtleties in data that commonly-used \textit{MLR} does not.

\subsubsection{\textbf{NYC Sale Price}}

\begin{figure}[t]
    \centering
    \begin{subfigure}[t]{0.99\columnwidth}
        \centering
        \includegraphics[width=\columnwidth]{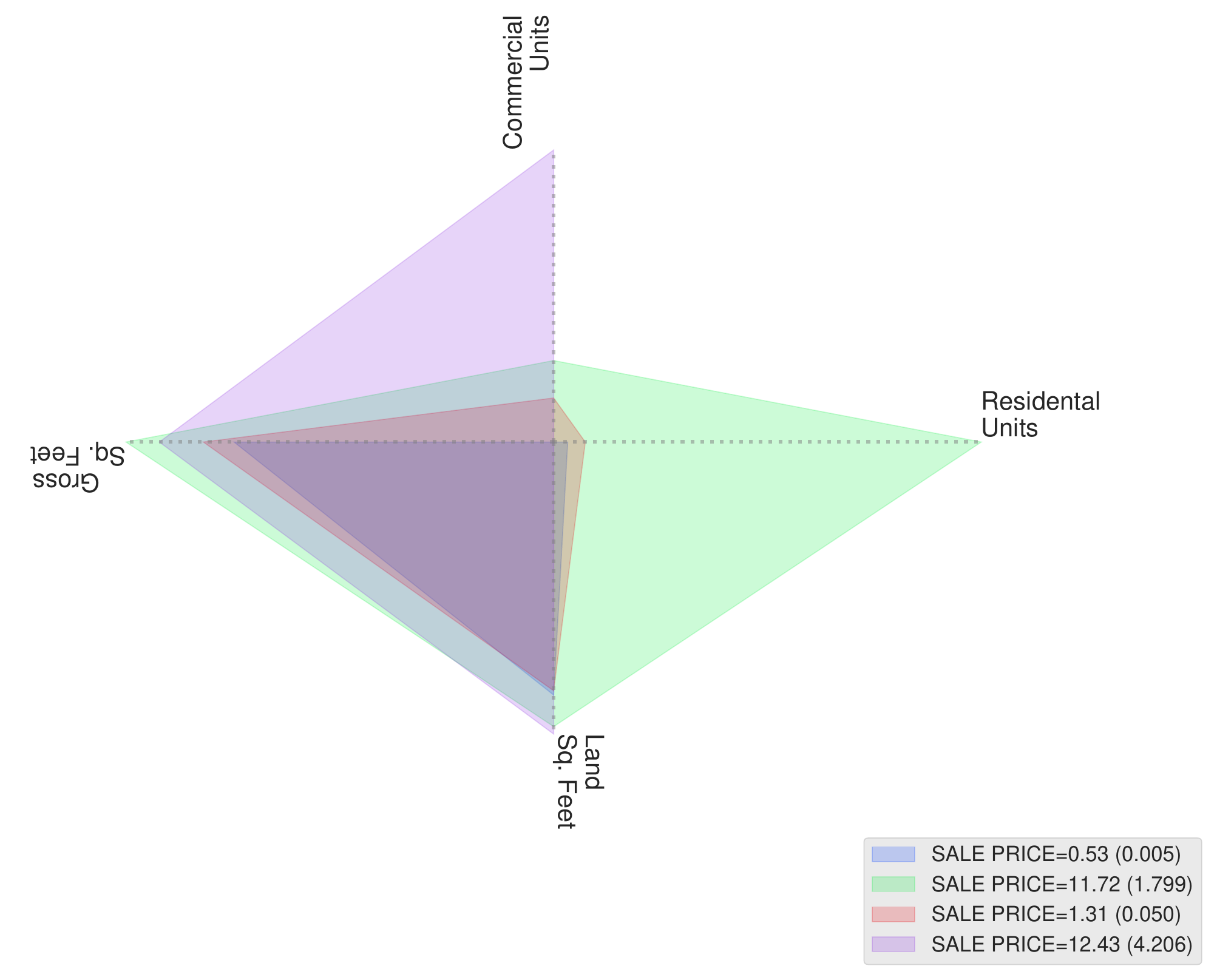} 
    \end{subfigure}
    
\resizebox{0.47\textwidth}{!}{
    \begin{tabular}{|c||c|c||l|l|l|l|}
        \hline
         & 
        \multicolumn{2}{c|}{\textit{Sale }} &  {\textit{Resid.}} & {\textit{Commer. }} & {\textit{Gross }} & {\textit{Land }} \\
         & 
        \multicolumn{2}{c|}{\textit{Price}} &  {\textit{Units}} & {\textit{Units}} & {\textit{Sq.Feet}} & {\textit{Sq.Feet}} \\
        \cline{2-7}
         & \textit{mean} & \textit{std err} & {$\beta$} & {$\beta$} & {$\beta$} & {$\beta$} \\
        \hline
        \textit{All} & $1.31$ & $0.079$ & $0.036$ & $0.039$ & $2.681$ & $0.595$ \\ 
        \hline
        \textit{Purple}  & $12.43$ & $2.143$ & $0.000$ & $0.218$ & $8.842$ & $-1.956$ \\
        \textit{Green} & $11.72$ & $0.916$ & $0.078$ & $1.008$ & $5.329$ & $-6.652$ \\
        \textit{Red} & $1.31$ & $0.026$ & $-0.142$ & $-0.900$ & $0.591$ & $-0.348$ \\
        \textit{Blue} & $0.53$ & $0.004$ & $-0.039$ & $0.000$ & $0.124$ & $0.089$ \\
        \hline
        \multicolumn{7}{|l|}{\footnotesize{all significant: p-value $\le 0.001$}} \\ \hline
    \end{tabular}}
    \caption{\label{fig:nyc_mu} The value of center ($\mu$) for four components of \textit{NYC} data. The numbers in legends are average price of the component's real estate and is in millions of dollars. 
    }
\end{figure}


The mean sale price of all properties in \textit{NYC} property sales data is \$1.31 Million. The mean price, however, hides a large heterogeneity in the properties on the market. Our method tamed some of this heterogeneity by identifying four components within the sales data. Figure~\ref{fig:nyc_mu} shows that these  components represent large commercial properties (\textit{Purple}), large residential properties (\textit{Green}),  mixed commercial/residential sales (\textit{Red}), and single-unit residential properties (\textit{Blue}). 

\begin{table}[h!]
    \centering
    
\resizebox{0.47\textwidth}{!}{
    \begin{tabular}{c|c|c|c|c|c}
        \textit{Component}& Manhattan & Bronx & Brooklyn & Queens & Staten \\ 
        \hline
            \textit{Purple} & \textbf{17}\% & 20\% & \textbf{41}\% & 17\% & 5\% \\
            \textit{Green} & \textbf{41}\% & {22}\% & \textbf{26}\% & 10\% & 1\% \\
            \textit{Red} & 8\% & 9\% & \textbf{65}\% & 15\% & 3\% \\
            \textit{Blue} & 0\% & 13\% & \textbf{37}\% & \textbf{33}\% & {16}\% \\
        \hline
        All  & 3\% & 13\% & 40\% & 30\% & 14\% \\ \hline
    \end{tabular}}
    \caption{DoGR components, and the boroughs that make up each component (rows might not add up to 100\% due to rounding).\label{tbl-NYC-boroughs}}
\end{table}

Table~\ref{tbl-NYC-boroughs} show what percentage of each component is made up of New York City's five boroughs. Large commercial properties (\textit{Purple} component), such as office buildings, are located in Brooklyn and Manhattan, for example. These are the most expensive among all properties, with average price of more than \$12 Million. The next most expensive type of property are large residential buildings (\textit{Green} component)---multi-unit apartment buildings. These are also most likely to be located in Manhattan and Brooklyn. Small residential properties (\textit{Blue} component)---most likely to be single family homes---are the least expensive, on average half a million dollars, and most likely to be located in Brooklyn and Queens, with some in Staten Island.

Regressions in this data set show several instances of Simpson's paradoxes. Property price in population-level data increases as a function of the number of \textit{residential units}. In disaggregated data, however, this is true only for the \textit{Green} component, representing apartment buildings. Having more residential units in smaller residential properties (\textit{Red} and \textit{Blue} components) lowers their price. This could be explained by smaller multi-unit buildings, such as duplexes and row houses, being located in poorer residential areas, which lowers their price compared to single family homes. Another notable trend reversal occurs when regressing on lot size (\textit{land square feet}). As expected, there is a positive relationship in the \textit{Blue} component with respect to lot size, as single family homes build on bigger lots are expected to fetch higher prices in New York City area. However, the trends in the other components are negative. This could be explained the following way. As land becomes more expensive, builders are incentivized to build up, creating multi-story  apartment and office buildings. The more expensive the land, the taller the building they build. This is confirmed by the positive relationship with \textit{gross square feet}, which are strongest for the \textit{Purple} and \textit{Green} components. In plain words, these components represent the tall buildings with small footprint that one often sees in New York City.

\subsubsection{\textbf{Stack Overflow}}
As our last illustration, we apply {\ourMethod} to Stack Overflow data to answer the question how well we can predict the length of the answer a user writes, given the features of the user and the answer.

\begin{figure}[h]
    \centering
\footnotesize{

\resizebox{0.47\textwidth}{!}{
\begin{tabular}{c|c|c|c|c|c}
\toprule
    & Size & Acceptance & Answerer & Num. of & Percentile \\ 
    & & Probability & Reputation & Answers & Tenure\\
    \midrule

    \textit{Red} & 39\% & 0.26 & 338.17 & 16.02 & 0.43 \\
    \textit{Green}& 47\% & 0.24 & 314.35  & 14.56 & 0.42 \\
            
    \textit{Blue} & 8\% & 0.33  & 492.19 & 20.65 & 0.47 \\
            
    \textit{Orange} & 5\% & 0.32  & 784.62  & 40.21 & 0.40 \\
    \midrule
    \end{tabular}
    }}
   \begin{subfigure}[t]{0.99\columnwidth}
        \centering
        \includegraphics[width=\columnwidth]{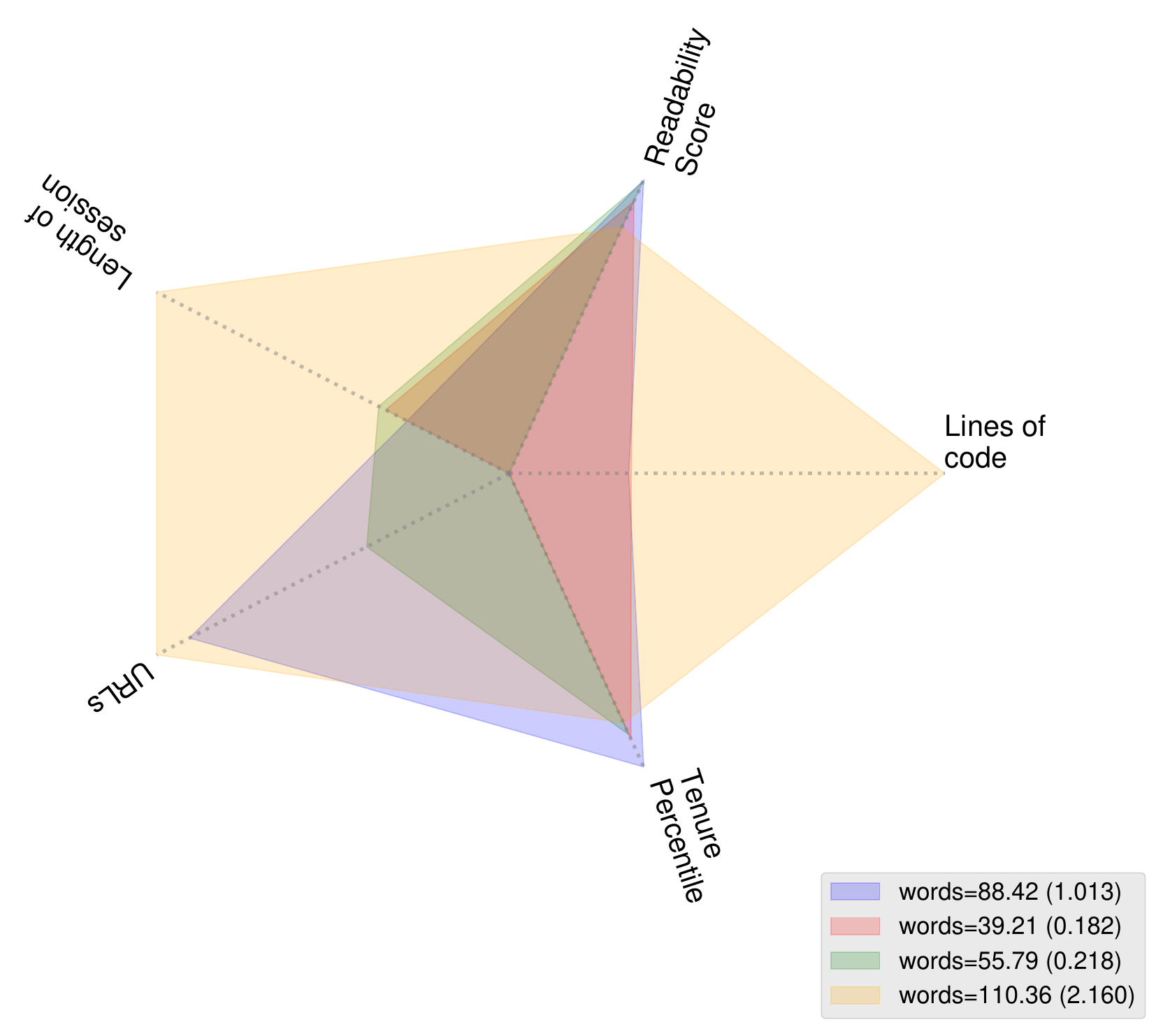} 
    \end{subfigure}
\resizebox{0.47\textwidth}{!}{
    \begin{tabular}{@{}c||c|c||l|l|l|l@{}}
        \hline
         & 
        \multicolumn{2}{c|}{\textit{Words}} &  {\textit{Codes}} & {\textit{Readab.}} & {\textit{Percnt.}} & {\textit{URLs}} \\
        \cline{2-7}
         & \textit{mean} & \textit{std} & {$\beta$} & {$\beta$} & {$\beta$} & {$\beta$} \\
        \hline
        \textit{All} & $54.91$ & $0.102$ & $0.29$ & $0.048$ & $5.61$ & $17.11$ \\ 
        \hline
        \textit{Red} & $39.21$ & $0.093$ & $0.43$ & $0.112$ & $6.09$ & $0$ \\
        \textit{Green} & $55.79$ & $0.111$ & $0$ & $-0.175$ & $2.64$ & $3.62$ \\
        \textit{Blue}  & $88.42$ & $0.517$ & $0.17$ & $-0.467$ & \emph{-0.22} & $-86.20$ \\
        \textit{Orange} & $110.36$ & $1.102$ & $-0.20$ & $0.341$ & $20.45$ & $12.87$ \\
        \hline
        \multicolumn{7}{l}{\footnotesize{\textit{* Italicized effects are not significant}}} \\ \hline
    \end{tabular}}
    \caption{\label{fig:se_fig} 
    Disaggregation of Stack Overflow data into subgroups. The outcome variable is length of the answer (number of words). The radar plot shows the importance of each features used in the disaggregation. The top table 
    shows average values of validation features, while the bottom table 
    shows regression coefficients for the groups.
    }
\end{figure}

{\ourMethod} splits the data into four clusters, as shown in \ref{fig:se_fig}. 
\textit{Green} and \textit{Red} components contain most of the data, with 47\% and 39\% of records respectively. The radar plot shows the relative strength of the features in each component, while the table above the plot shows features characterizing each discovered group. Except for \textit{Percentile tenure}, these features were not used by {\ourMethod} and are shown to validate the discovered groups. 

The \textit{Orange} component ($5\%$ of data) contains very active (longer \textit{Session Length}) users, who meticulously document their answers with many \textit{lines of code} and \textit{URLs}, so we can label them ``power users''. These are among the longest (high \textit{Words}) and more complex (low \textit{Readability}) answers in the data, and they also tend to be high quality (high \textit{Acceptance Probability}). Surprisingly this group has newer users (lower \textit{Percentile Tenure}), but they have high reputation (\textit{Answerer Reputation}) and wrote more answers previously (higher \textit{Number of Answers}). These users, while a minority, give life to Stack Overflow and make it useful for others. 
\textit{Orange} component users have more \textit{code lines} within shorter answers. This is in contrast to other groups, which tend to include more lines of code within longer answers. The brevity of \textit{Orange} users when documenting code is an example of a trend reversal.

Another interesting subgroup is the \textit{Blue} group ($8\%$ of data), which is composed of ``veterans'' (high \textit{Percentile Tenure}), who write easy-to-read answers (high \textit{Readability}) that are documented with many \textit{URLs}. These users have a relatively high \textit{reputation}, but they are selective in the questions they answer (lower \textit{Number of Answers} than for the \textit{Orange} users). Interestingly, tenure (\textit{Percentile}) does not have an effect on length of the answer for these \textit{Blue} users, while it has positive effect in other groups (i.e., more veteran users write longer answers). Negative effect of \textit{URLs} on the length of the answer suggests that these users use \textit{URL}s to refer to existing answers. 

The \textit{Red} (39\%) and \textit{Green} (47\%) components contain the vast majority of data. They are similar in terms of user \textit{reputation},  tenure (\textit{Percentile}) and experience (\textit{Number of Answers}), as well as the quality of the answers they write (\textit{Acceptance Probability}), with the \textit{Red} component users scoring slightly higher on all the measures. The main difference is in the answers they write: \textit{Red} users do not include \textit{URLs} in their answers, while \textit{Green} users do not include \textit{code}. 
Another difference between these groups is that \textit{Green} users have longer answers than \textit{Red} users, but as their answers become longer, they also become more difficult to read (lower \textit{Readability}). In contrast, longer answers by  \textit{Red} users are easier to read.

Overall, we find intuitive and surprising results from data that are largely due to DoGR's interpretability.

\subsection{Quantitative Results} 
We compare the performance of {\ourMethod} to existing state-of-the-art methods for disaggregated regression: the three variants of CLR, 
WCLR~\cite{silva2017}, FWCLR~\cite{silva2018}, 
and GMR~\cite{sung2004}. We use \textit{CART} as a method which uses decision tree for regression. We also use \textit{MLR}, which does not disaggregate data, as a baseline.

\subsubsection{Prediction Performance}
For the prediction task, we  use $5\times5$-fold nested cross validation to train the model on four folds and make predictions on the out-of-sample data in the fifth fold. As hyperparameters, we used $k =$ 1--6 as potential number of components for all methods. For \textit{WCLR and FCWCLR}, we set $\alpha = \{ 0.001, 0.01, 0.1, 10, 100, 1000 \}$, and for \textit{FWCLR} we set $m = \{1.1, 1.5, 2.0, 3.0\}$ for \textit{CART} we set the depth of the tree as $\{ 1, 2, ..., 9\}$. 
We use grid search to find best hyperparameters. Table \ref{tab:pred} presents results on our datasets and synthetic data. 
To evaluate prediction quality, we use \textit{Root Mean Square Error (RMSE)} and \textit{Mean Absolute Error (MAE)}. When comparing the quality of cross-validated predictions across methods, we use the Kruskal-Wallis test, a non-parametric method to compare whether multiple distributions are different \cite{Kruskal1952}. 
When distributions were significantly different, a pairwise comparisons using the Tukey-Kramer test was also done as post-hoc test, to determine if mean cross-validated quality metrics were statistically different \cite{McDonald2014}. 
Prediction results are shown in table \ref{tab:pred}. We use $*$ to indicate statistically significant ($p<0.05$) differences between our method and other methods. Bolded values differ significantly from others or have lower standard deviation. Following \cite{silva2018}, we use lower deviation to denote more consistent predictions.

\textit{CART}, \textit{WCLR} and \textit{FWCLR} do not perform well in \textbf{Synthetic} data, because variances of the two components are different, which these methods do not handle. 
This is a problem that exists in an arbitrary number dimensions; we observe in higher dimensions the gap between performance of \textit{FWCLR} and our method \textit{increases}. 

\begin{figure}
    \centering
    \begin{subfigure}{.23\textwidth}
        \includegraphics[width=\textwidth]{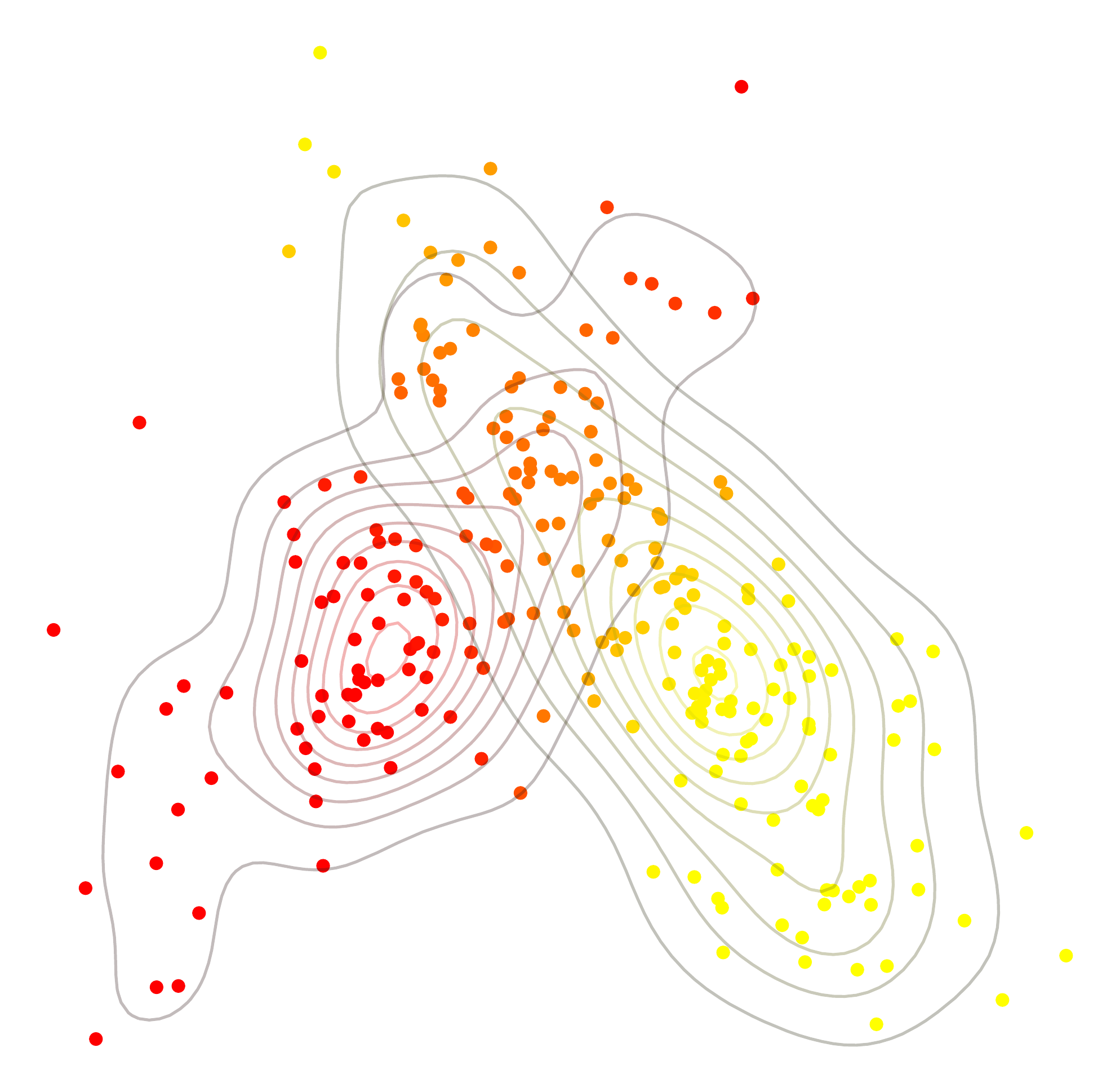}
        \caption{Soft Clustering}\label{fig:soft_clus}
    \end{subfigure}
    \begin{subfigure}{.23\textwidth}
        \includegraphics[width=\textwidth]{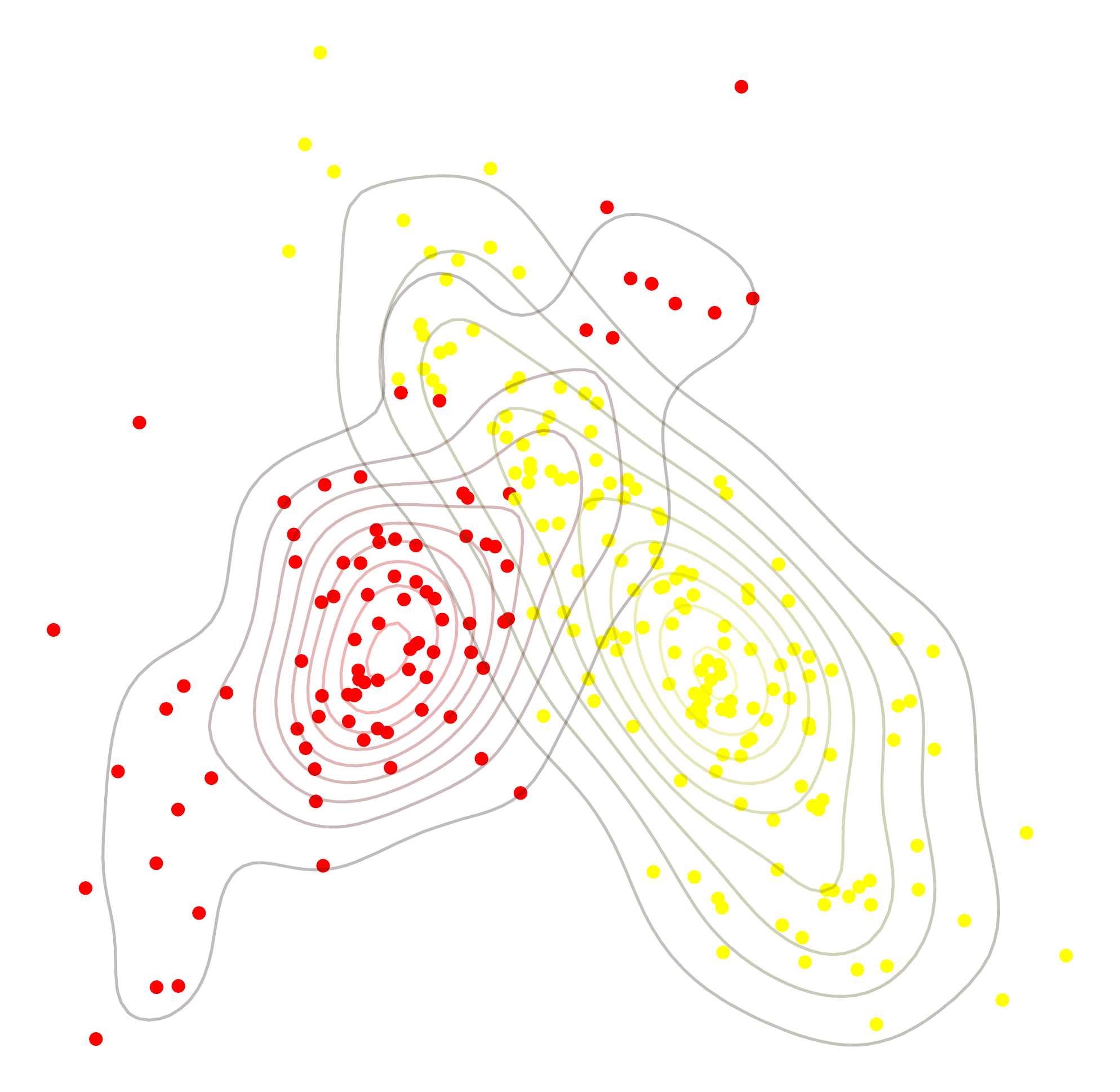}
        \caption{Hard Clustering}\label{fig:hard_clus}
    \end{subfigure}
    \caption{The difference between hard and soft clustering for data analysis. Assuming we have two clusters (yellow and red); in hard clustering, the uncertainty of cluster assignment is not tractable in data analysis phase (e.g. studying the coefficients of the independent variables), since all the data-points have one of the main cluster colors (b). However, having the soft-clustering, we can get the approximated coefficient for each individual data-point (the whole range from yellow to red in (a)), separately.}
    \label{fig:hard_vs_soft}
\end{figure}

For \textbf{Metropolitan} data, there is no statistically significant difference between methods: all cluster-based regression methods outperform \textit{MLR} and \textit{CART} on the prediction task. 
This shows that any method that accounts for the latent structure of data helps improve predictions. 
For \textbf{Wine Quality} data, while the null hypothesis of equal performance of \textit{GMR} and {\ourMethod} cannot be rejected, our method has a smaller standard deviation in both datasets.
We were not able to successfully run \textit{GMR} and \textit{WCLR} on \textbf{NYC} data, due to exceptions (returning null as predicted outcome) and extremely long run time, respectively. It took three days for \textit{FWCLR}, and four hours for our method to finish running on \textit{NYC} data. Our method has significantly lower \textit{MAE} compared to \textit{FWCLR}, while \textit{CART} has a better \textit{MAE} and worse \textit{RMSE}. This shows that \textit{CART} has a worse performance for outliers. Figure \ref{fig:hard_vs_soft} represents the reason behind the poor performance for outliers for hard clustering methods like \textit{CART}.

We were also not able to successfully run \textit{GMR} and \textit{WCLR} on \textbf{Stack Overflow} data for the same reasons. 
It took six days for \textit{FWCLR} to run one round of cross-validation, after which we stopped it. Therefore, the mean reported in the table for \textit{FWCLR} is the average of five runs, while for \textit{MLR} and {\ourMethod} it is the average of $25$ runs. The best performing method is \textit{CART}. The main reason is that \textit{Stack Overflow} has discrete variables and \textit{CART} is a better method is compare with \textit{FWCLR} and {\ourMethod} for handling that types of variables. 

\begin{table}[t!]
    \centering
\resizebox{0.45\textwidth}{!}{
    \begin{tabular}{c|c|c}
        Method & RMSE ($\pm \sigma$) & MAE ($\pm \sigma$) \\
        
        \hline 
        \multicolumn{3}{c}{Synthetic} \\ 
        \hline
        MLR & 294.88 ($\pm$ 1.236)* & 288.35 ($\pm$ 0.903)* \\ 
        
        CART & 264.70 ($\pm$ 7.635)* & 224.57 ($\pm$ 6.138)* \\ 
        WCLR & 261.14 ($\pm$ 3.370)* & 232.76 ($\pm$ 2.682)* \\ 
        
        FWCLR & 261.27 ($\pm$ 4.729)* & 233.05 ($\pm$ 3.772)* \\
        
        GMR & 257.36 ($\pm$ 4.334) & 219.15 ($\pm$ 3.567) \\ 
        
      \textbf{\ourMethod} & \textbf{257.32 ($\pm$ 3.871)} & \textbf{219.11 ($\pm$ 3.106)} \\ 

        \hline 
        \multicolumn{3}{c}{Metropolitan} \\ 
        \hline
        
        MLR & 0.083 ($\pm$ 0.0061) & 
        0.062 ($\pm$ 0.0033) \\ 
         
        CART & 0.086 ($\pm$ 0.0056) & 0.064 ($\pm$ 0.0036)* \\  
        WCLR & 0.083 ($\pm$ 0.0029) & 0.062 ($\pm$ 0.0024)\\ 
        
        FWCLR & \textbf{0.082 ($\pm$ 0.0044)} & \textbf{0.061 ($\pm$ 0.0021)} \\
        
        GMR & 0.083 ($\pm$ 0.0043) & 0.061 ($\pm$ 0.0023) \\ 
        
      \textbf{\ourMethod} & 0.083 ($\pm$ 0.0052) & 
      0.061 ($\pm$ 0.0031) \\ 
    
        \hline 
        \multicolumn{3}{c}{Wine Quality} \\ 
        \hline
        MLR & 0.83 ($\pm$ 0.018)* & 0.64 ($\pm$ 0.015)* \\ 
        
        CART & 0.79 ($\pm$ 0.015) & 0.62 ($\pm$ 0.013) \\ 
        WCLR & 0.83 ($\pm$ 0.013)* & 0.64 ($\pm$ 0.011)* \\
        FWCLR & 0.80 ($\pm$ 0.013)* & 0.63 ($\pm$ 0.009)* \\
          
        GMR & 0.79 ($\pm$ 0.017) & 0.62 ($\pm$ 0.014) \\ 
       \textbf{\ourMethod} & \textbf{0.79 ($\pm$ 0.014)} & \textbf{0.62 ($\pm$ 0.011)} \\ 
        
        \hline
        \multicolumn{3}{c}{NYC} \\ 
        \hline
        MLR & 13.36 ($\pm$ 7.850) & 2.20 ($\pm$ 0.064)* \\ 
        CART & 15.33 ($\pm$ 9.128) & \textbf{1.34 ($\pm$ 0.190)} \\ 
        FWCLR & 13.14 ($\pm$ 7.643) & 1.76 ($\pm$ 0.321)* \\
       \textbf{\ourMethod} & \textbf{11.88 ($\pm$ 9.109)} & 1.40 ($\pm$ 0.222) \\ 

        \hline
        \multicolumn{3}{c}{Stack Overflow} \\ 
        \hline
        MLR & 60.69 ($\pm$ 1.118) & 37.74 ($\pm$ 0.152) \\ 
        
        CART &  \textbf{58.19 ($\pm$ 0.781)}* & \textbf{34.05 ($\pm$ 0.208)}* \\ 
        
        FWCLR & 60.47 ($\pm$ 0.960) & 37.25 ($\pm$ 0.794) \\ 
        
      \textbf{\ourMethod} & 60.68 ($\pm$ 1.298) & 37.62 ($\pm$ 0.314) \\ 
        
    \end{tabular}}
    \caption{Results of prediction on five data-sets. 
    Asterisk $*$ indicates results that are significantly different from our method (p-value $< 0.05$). The bolded results have smaller standard deviation among the best methods with same mean of error. 
    \label{tab:pred}}
\end{table}

\subsubsection{\textbf{Run Time}}
To compare the run time of all algorithms, we performed one round of cross validation (not nested) for each method. The same machine ($4$-GB RAM, $3.0$-GHz Intel CPU, Windows OS) was used for time calculation. The available code for \textit{WCLR} and \textit{FWCLR} methods are in \textit{R} while the other methods are written in \textit{Python}. Table \ref{tab:time} presents the run time in minutes. The slowest method is \textit{WCLR}, while the fastest one is \textit{MLR}. \textit{WCLR} and \textit{FWCLR} are sensitive to size of data, 
perhaps due to the many hyperparameters they need to tune. To find the best hyperparameters for \textit{GMR} and {\ourMethod}, we ran the methods $6$ times, while \textit{WCLR}  requires $36$ runs and  \textit{FWCLR} $144$ runs. 
Beside \textit{NYC} and \textit{Stack Overflow} datasets for which exceptions occur in \textit{GMR}, the run time of {\ourMethod} is twice that of \textit{GMR}. The reason \textit{GMR} throws exception could be because of a singular covariance matrix.

While our method is similar in spirit to \textit{GMR}, 
out method is more stable, 
as shown on the \textit{NYC} and \textit{Stack Overflow} data. In addition, our method is interpretable, as it directly computes regression coefficients, while \textit{GMR} represents relationships between variables via the covariance matrix. 
Covariance values are not guaranteed to have the same sign, let alone magnitude, as regression coefficients. Regression is therefore necessary to understand the true relationships between variables. Mathematically, \textit{GMR} and \textit{unweighted} regression can be converted to one another using linear algebra. It is not clear, however, whether the equivalence also holds for weighted regression. 

\begin{table}[h!]
    \centering
\resizebox{0.40\textwidth}{!}{
    \begin{tabular}{|c|c|c|c|c|}
    \hline
    Dataset & WCLR & FWCLR & GMR & \ourMethod \\ 
    \hline
    Synthetic & 6.46 & 0.68 & \textbf{0.76} &  1.4\\ 
    Metropolitan & 37 & 9 & \textbf{2} & 4\\
    Wine Quality & 180+ & 36 & \textbf{9} & 16\\
    NYC & 600+ & 232 & failed & \textbf{10} \\ 
    Stack Overflow & - & 7200+ & failed & \textbf{170} \\ 
    \hline
    \end{tabular}}
    \caption{Run time (in \textit{minutes}) of one round of cross-validation (non-nested). \textit{MLR} took less than a minute for all datasets. 
    'failed' indicates that the method could not run on the data due to exceptions. We time out WCLR exceeding sufficient amount of time to show the order of performance. \textbf{Bold} numbers indicate the fastest algorithm.
    \label{tab:time}}
\end{table}

\section{Discussion}

In this paper, we introduce {\ourMethod}, which softly disaggregates data by latent confounders. 
Our method retains the advantages of linear models, namely their interpretability, by reporting regression coefficients that give meaning to trends. Our method also discovers the multidimensional latent structure of data by partitioning it into subgroups with similar characteristics and behaviors. While alternative methods exist for disaggregating data, our approach produces interpretable regressions that are computationally efficient.




We demonstrated the utility of our approach by applying it to real-world data, from wine attributes to answers in a question-answering website. We show that our method identifies meaningful subgroups and trends, while also yielding new insights into the data. For example, in the wine data set {\ourMethod} correctly separates high quality red wines from white, and also  discovers two distinct classes of high quality white wines. In Stack Overflow data, {\ourMethod} identifies important users like ``veterans'' and ``power users.''

There are a few ways to further improve our method. Currently, it is applied to continuous variables, but it needs to be extended to categorical variables, often seen in social science data. Moreover real data is often non-linear; 
therefore, our method needs to be extended to non-linear models beyond linear regression. In addition, to improve prediction accuracy, our method could include regularization parameters, such as ridge regression or LASSO. 
Even in its current form, however, {\ourMethod} can yield new insights into data.



\small
\bibliography{main}
\bibliographystyle{aaai}

\end{document}